\documentclass{article}
\usepackage{ijcai22}

\usepackage{times}

\usepackage{soul}
\usepackage{url}
\usepackage[hidelinks]{hyperref}
\usepackage[utf8]{inputenc}
\usepackage[small]{caption}
\usepackage{graphicx}
\usepackage{amsmath}
\usepackage{booktabs}
\usepackage{multirow}
\usepackage{caption}
\usepackage{float}
\usepackage[toc,page]{appendix}
\title{Evaluating the Faithfulness of Saliency-based Explanations for Deep Learning Models for Temporal Colour Constancy}

\author{
Matteo Rizzo$^1$\footnote{Contact Author}\and
Cristina Conati$^1$\and
Daesik Jang$^2$\and
Hui	Hu$^2$\\
\affiliations
$^1$ University of British Columbia
$^2$ Huawei\\
\emails
$^1$ matteo.rizzo.phd@gmail.com conati@cs.ubc.ca
$^2$ \{daesik.jang, huhui12\}@huawei.com
}

\begin{document}

\maketitle

\begin{abstract}
The opacity of deep learning models constrains their debugging and improvement. Augmenting deep models with saliency-based strategies, such as attention, has been claimed to help get a better understanding of the decision-making process of black-box models. However, some recent works challenged saliency’s faithfulness in the field of Natural Language Processing (NLP), questioning attention weights’ adherence to the true decision-making process of the model. We add to this discussion by evaluating the faithfulness of in-model saliency applied to a video processing task for the first time, namely, temporal colour constancy. We perform the evaluation by adapting to our target task two tests for faithfulness from recent NLP literature, whose methodology we refine as part of our contributions. We show that attention fails to achieve faithfulness, while confidence, a particular type of in-model visual saliency, succeeds. 
\end{abstract}

\section{Introduction}
\label{sec:introduction}

Despite achieving impressive accuracy in countless tasks, Deep Learning (DL) models are inherently black-boxes, and thus their inner decision-making process is hard for humans to understand. Such opacity constraints the evaluation of their generalization power and may hinder model debugging and improvement.

The long-term goal of our research is to investigate how to increase the interpretability of DL models, specifically focusing on models for Sequential Data (SD). Existing research on the interpretability of DL models for SD has largely concentrated on language data \cite{danilevsky-etal-2020-survey}, with preliminary results that are mostly anecdotal (e.g., \cite{jain2019attention,wiegreffe2019attention,serrano2019is}). Research on the interpretability of DL models for non-SD is more advanced for well-known tasks such as image object recognition \cite{hohman2019visual} but it is not clear if and how existing results transfer to other models and tasks. In particular, despite the growing interest in video-related tasks brought about by increasingly more powerful hardware, few works addressed the interpretability of models operating on video data (see \cite{DBLP:journals/corr/abs-1909-05667} for a survey).

\begin{figure}[t]
    \centering
    \includegraphics[width=0.45\textwidth]{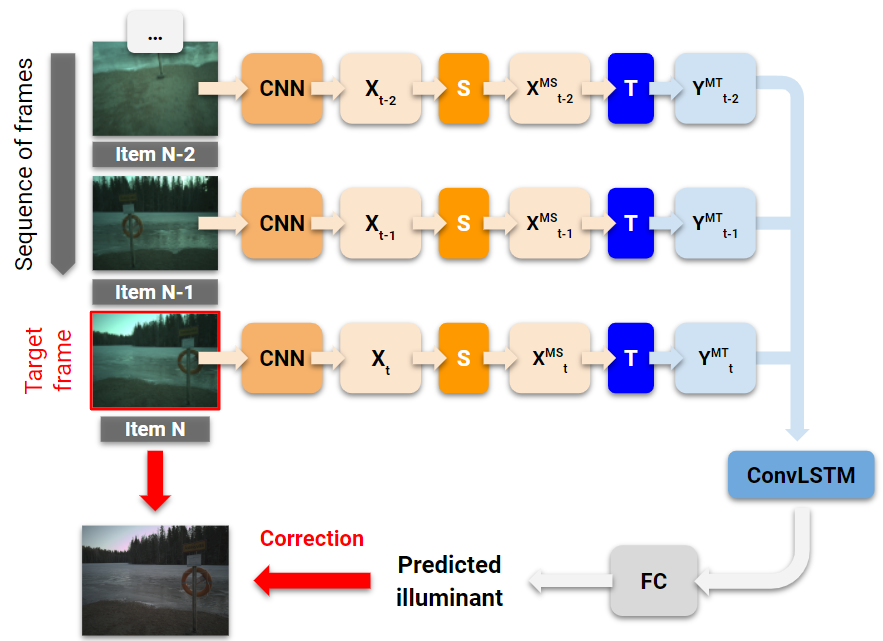}
    \caption{Example of CNN+LSTM architecture for the TCC task.}
    \label{fig:tcc-architecture}
\end{figure}

We aim to extend these works by looking at the interpretability of DL models for a task known as computational colour constancy for video data (Temporal Colour Constancy, or TCC). The TCC task aims to recursively estimate the illuminant colour of each frame in a video sequence accounting for a window of preceding frames \cite{qian2020a}. TCC methods find prominent application in the processing pipeline of digital cameras, to increase the output video quality by compensating for colour distortions \cite{ramanath2005color}. An example is shown in the left side of Figure \ref{fig:tcc-architecture}, where the green illuminant in the target frame (i.e., the last in the sequence) is detected using information from the previous frames and used to perform the colour correction. The State-of-the-Art (SoA) approaches for TCC are DL-based, and the top-performing methods employ a CNN+LSTM architecture known as TCCNet \cite{qian2020a}, but so far there is limited work on understanding when these models work well, when they don’t, and why. 

We investigate how to increase the interpretability of CNN+LSTM models for TCC to provide model developers, our target users, with useful insights into the inner workings of the models, such as which features are the most relevant to the model prediction, to support model debugging and improvement. On a practical level, one way to answer such a question is to point the stakeholders to salient parts of the input sequence. To address this problem, we look at in-model saliency techniques, such as attention \cite{bahdanau2015neural}, where in-model means that the technique modifies the model architecture itself. In-model saliency methods (referred to just as saliency methods from now on) have been initially proposed to improve model accuracy, but they have also been considered very promising to increase interpretability. These methods consist of adding to the model architecture a mechanism (often in the form of one or more additional layers) to learn a set of weights associated with each encoded input component (e.g., each word in a text, pixel in an image, frame in a video, etc.), guiding the model to focus on specific input components to predict the output. The idea is that the weights identify which input components are more important for prediction, and therefore visualizing them in saliency-maps, such as heatmaps, provides explanations on the relevance of input components for prediction

For instance, attention \cite{bahdanau2015neural} has pushed the SoA accuracy of multiple tasks for both Natural Language Processing (NLP) \cite{correia2021attention} and computer vision \cite{https://doi.org/10.48550/arxiv.2111.07624}, and heatmaps of the attention weights have been leveraged as explanations to provide a sense of what parts of the original input were the most discriminatory for the model, (e.g., \cite{choi2016retain,lei2017interpretable,mullenbach2018explainable}).

Attention can be implemented either as a standalone module or be learned together with other feature maps. In the latter case, we speak of confidence, a concept introduced in the FC4 method for single-frame computational colour constancy \cite{hu2017fc} but never further investigated, neither for increasing accuracy nor interpretability. It is interesting to look at confidence as its weights, unlike in attention,  are jointly learned with the other feature maps and thus might have a stronger relationship with the inner reasoning of the model.

Our long-term goal is to explore both attention and confidence as means to increase the interpretability of the CNN+LSTM architecture which represents the SoA approach for tackling TCC \cite{qian2020a}. In this paper, we focus on evaluating two aspects that are critical to ascertain the suitability of an in-model interpretability technique. The first is verifying that modifications made to the original model architecture do not reduce model accuracy. The second aspect relates to making sure that the technique is \textit{faithful}, namely it generates explanations that accurately represent the decision-making processes of a model \cite{jacovi-goldberg-2020-towards,jacovi2021}.  Considering faithfulness is important because some researchers started questioning whether saliency maps of attention weights provide accurate insights into how the model learns to decode the weighted encodings into a prediction \cite{jain2019attention,wiegreffe2019attention,serrano2019is,10.5555/3327546.3327621}. These works introduced tests that evaluate faithfulness from different perspectives, with results that ultimately fail to provide conclusive evidence about whether attention should or should not be relied upon when devising explanations for a DL model. Hence, in this paper, we extend two of these tests to our TCC task to ascertain if attention and confidence can be considered faithful interpretability techniques for this task. 

For each of these two saliency techniques and the combination of the two, we apply them to the spatial components of our architecture (CNN), the temporal component (LSTM) and both, thus obtaining nine models that we evaluate for accuracy and faithfulness. In terms of accuracy and interpretability, the spatial and temporal dimensions of in-model saliency have not been extensively compared, thus our evaluation aims to contribute to filling this gap. Our contributions are the following:

\begin{itemize}
    \item We dive into the largely uncharted territory of interpretability of in-model saliency explanations for video data, experimenting with the under-explored task of illuminant estimation from sequences of frames (i.e., TCC).
    \item We analyze the faithfulness of two different types of saliency (i.e., attention and confidence) and three different dimensions of saliency (i.e., spatial, temporal, and both combined). A formal comparison among these different types and dimensions of saliency is under-investigated in current literature.
    \item We extend tests for faithfulness previously investigated only for NLP to the TCC scenario by applying a rigorous methodology.
\end{itemize}

Our code is publicly available at: https://github.com/matteo-rizzo/saliency-faithfulness-eval

\section{Related Work}
\label{sec:related-work}

Attention has been leveraged in video processing mostly to improve accuracy in tasks other than TCC, such as Action Recognition \cite{sharma2016action,quan2019attention,li2018videolstm,song2017an,meng2019interpretable}, Multi-label Topic Classification \cite{kim2018temporal,you2020attention}, Captioning \cite{xu2017learning,yan2020stat,zhu2019attention} and Question Answering \cite{zhao2017video}. Only Meng et al. started looking at attention for interpretability in videos, namely for the Action Recognition task \cite{meng2019interpretable}. They propose a basic yet highly decoupled and easy plug-in spatiotemporal attention mechanism and show quantitative evidence that this can successfully isolate action-related content both spatially and temporally, testing it against two datasets for spatial and temporal action localization respectively. As a baseline strategy, we adopt the attention mechanism proposed by Meng et al. to see if their findings generalize to our very different TCC task. 

Thus far only a couple of preliminary investigations have applied in-model saliency in the form of attention to computational colour constancy \cite{yan2020stat,zhang2020adcc}, both for single images, not videos. Yan et al. detected a small accuracy boost using channel-wise attention on top of a fully convolutional network but did not examine interpretability implications \cite{yan2020stat}. Zhang et al. also reported improved accuracy when processing edge-augmented log-chrominance histograms with a self-attention DenseNet. They attempt attention visualizations to gain insights into the inner workings of their proposed method, but the discussion is limited to the scrutiny of two exemplary inputs \cite{zhang2020adcc}.

Another in-model saliency-based method we explore in this work is confidence. This was introduced as a particular form of visual attention in the FC4 method for single-frame computational colour constancy \cite{hu2017fc}. Whereas visual attention is generally learned by a standalone convolutional module, confidence is an additional feature map that weights the importance of the per-patch colour estimates generated by a fully convolutional backbone. Hu et al. report that confidence-based feature pooling leads to superior accuracy for illuminant estimation, and confidence scores showed potential for interpretability when plotted as a heatmap over the original input to highlight salient image patches.

Our evaluation of in-model saliency focuses on the property of faithfulness of explanation \cite{jacovi-goldberg-2020-towards,jacovi2021}. Despite the widespread interest in attention to increasing interpretability, only a few works attempted to evaluate the faithfulness of saliency-based explanations. Such evaluations concentrated on textual data \cite{serrano2019is,jain2019attention,wiegreffe2019attention} with only one investigation on visual data \cite{10.5555/3327546.3327621} and none involving video data. 

In this paper, we extend to video data two tests proposed by Wiegreffe \& Pinter \cite{wiegreffe2019attention} to evaluate the faithfulness of attention in NLP tasks. The first test (WP1 from now on) is designed to evaluate whether attention weights have a relevant impact on task accuracy in the first place. The second test (WP2 from now on) tries to gauge whether attention weights embed information about the relationship among input timesteps. Details on these tests are provided in section \ref{sec:original-methodology}.

Jain \& Wallace \cite{jain2019attention} propose two different tests, also to evaluate the faithfulness of attention in NLP tasks. One test involves a comparison with alternative measures of input feature importance, e.g. gradient-based measures, assuming that attention-based importance is faithful if the feature importance weights it generates highly correlates with those generated by the other measures. We do not look at this test because it relies on the unverified assumption that the alternative measures of feature importance are faithful. The second test involves analyzing whether replacing learned attention weights with different distributions affects model prediction, with the assumption that if this change does not affect prediction, then weights are not involved in the decision process and thus they cannot provide faithful explanations of such process. This test is complementary to the aforementioned WP1 in that it checks the importance of learning the weights through the attention mechanism, whereas WP1 checks that any weights have a role in the decision process in the first place. The tests in \cite{jain2019attention} are used to evaluate an LSTM architecture with attention, and the tasks of Binary Text Classification, Question Answering and Natural Language Inference on various datasets. Their results indicate that attention mostly failed the tests with this model and tasks, suggesting that this has limited reliability for interpretability.

The tests in Serrano \& Smith \cite{serrano2019is} aim to understand how well attention weights represent the importance of the encoded input components by zeroing out sets of weights and seeing how this affects prediction. They tested different versions of a Hierarchical Attention Network (HAN) on four large-scale multi-class text classification datasets. Their findings indicate that attention weights are poor indicators of the importance of encoded input components. However, their methodology is limited to looking at trends in plots and lacks a quantitative assessment of whether a model passed or failed. These are common issues across the existing evaluations of faithfulness \cite{jain2019attention,wiegreffe2019attention,serrano2019is}, which we address when applying the  WP1 and WP2 tests from  Wiegreffe \& Pinter \cite{wiegreffe2019attention} to the TCC task.

\section{Proposed Neural Architectures}
\label{sec:original-methodology}

To extensively evaluate saliency faithfulness for TCC, we trained a total of nine different models. We look at three dimensions of CNN+LSTM architecture, namely, Spatial (S), Temporal (T), and Spatio-Temporal (ST). Furthermore, we examine two types of saliency: attention (A) and confidence (C). Despite both being learned to be used to weight input components after the encoding phase, their implementation differs. It is interesting to look at confidence interpretability-wise as its tighter connection with the output prediction may foster more faithful explanations. As a simple exploratory form of integration of attention and confidence, we also examine the architecture using confidence as spatial saliency and attention as temporal saliency (CA). We leave the investigation of other (possibly more sophisticated) integrations for future work.

Figure \ref{fig:tcc-architecture} shows the basic CNN+LSTM backbone augmented with saliency. The spatial saliency module processes each CNN-encoded frame $X_i$ in the sequence individually, learning a mask $MS_i$. Then, the sequence of masked encoded frames $X^{MS}=X \cdot MS$ is fed to the ConvLSTM equipped with temporal saliency, which learns a temporal mask $MT_i$ contextually to each processed timestep. The output of this process is a series of temporally encoded timesteps $Y_i$ weighted as $Y^{MT}=X \cdot MT$. Finally, the output of the ConvLSTM is passed through a Fully Connected (FC) layer to get the illuminant prediction used to colour correct the last frame in the sequence.

The implementation of the attention modules, both spatial and temporal, was adapted from \cite{meng2019interpretable}, who originally devised them for a video action recognition task. Spatial attention is learned via a three-layers CNN module downscaling the feature maps to one channel. The first two layers use batch normalization and ReLU activations while the last one features a Sigmoid as activation and no normalization. As for temporal attention, this is learned as the Softmax of the output of two feed-forward neural networks which are jointly trained with all other components of the system. At each timestep, the T attention mechanism attends to every timestep in the sequence ($X_i^{MS}$) and the previous hidden state ($H_{t-1}$). The final feature map that is fed to the ConvLSTM is a weighted sum of the features from all the frames.

Confidence, instead, is a concept that naturally applies to the spatial dimension as presented in the original paper for single-frame computational colour constancy in a pooling strategy \cite{hu2017fc}. Spatial confidence is learned as an additional channel to the feature maps and used to weigh the encoded images. To derive temporal weights from spatial confidence masks we take the average value of each mask which, as shown in the supplementary material of \cite{hu2017fc}, positively correlates with the accuracy of prediction for single-frame computational colour constancy.

\begin{figure}[t]
    \centering
    \includegraphics[width=0.49\textwidth]{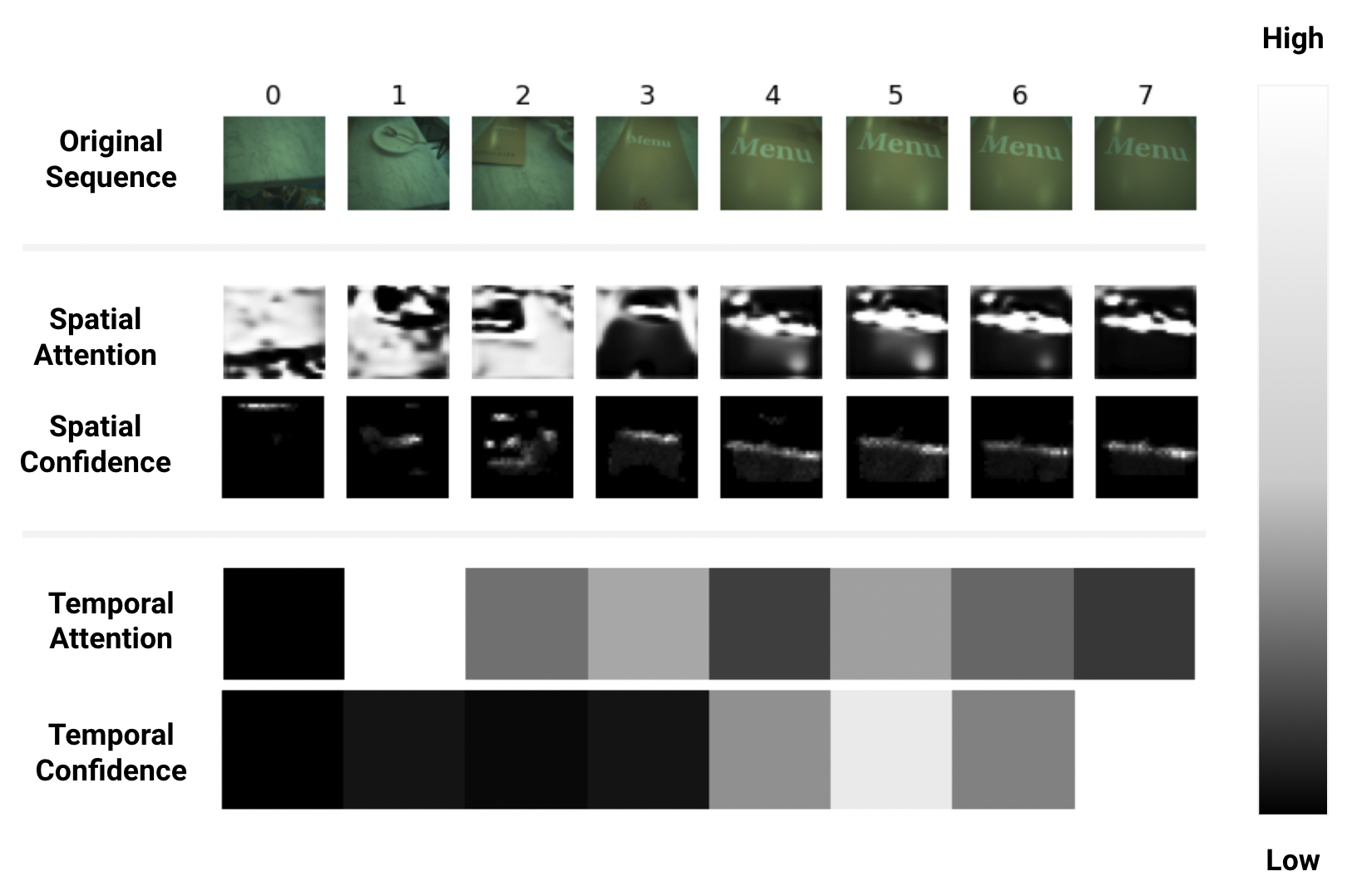}
    \caption{Saliency heatmaps for attention and confidence.}
    \label{fig:heatmaps}
\end{figure}

Conveniently, both attention and confidence can easily be visualized through heatmaps. In our TCC scenario, this can be done both spatially and temporally (Figure \ref{fig:heatmaps}). Intuitively, by comparing the heatmaps with the original frame in the sequence we get a sense of which input features contributed the most to the model prediction.

\section{Original Methodology of the Tests}
\label{sec:original-methodology}

Test WP1 aims at assessing whether the attention module has an impact on accuracy when applied to some deep neural architecture for a given task and dataset. The authors claim that saliency cannot be part of the decision-making process of a model if it is not used for generating more accurate predictions. Originally, the test was applied to a vanilla LSTM architecture. This is a Contextual (C) architecture, in that it embeds a layer, the recurrent one, capable of grasping temporal relationships among the input components. In a Non-Contextual (NC) architecture, on the other hand, C layers are replaced by linear layers.

WP1 requires inferring a prediction with a given contextual attention model $M^C$ twice: (i) using standard learned saliency weights ($M_C^C$) and (ii) using frozen random uniformly distributed weights ($M_U^C$)). If the accuracy achieved by $M_C^C$ outperforms that achieved by $M_U^C$ then the test is passed, and it is failed otherwise. More specifically, the authors claim that: (i) if $M_C^C$ outperforms $M_U^C$ then attention is actively involved in the decision making process of the model and thus should be looked at for faithfulness; (ii) if $M_C^C$ does not outperform $M_U^C$ then attention is not actively involved in the decision making process of the model and thus should not be further investigated for faithfulness. WP1 is a necessary condition for faithfulness and a precondition for WP2.

Intuitively, we would like saliency-augmented models to use saliency in their decision-making process to highlight important input components and filter out unimportant input components. For this purpose, saliency needs to encode contextual information. Test WP2 looks at whether learned saliency weights encode contextual information. More specifically, it looks at whether replacing the weights learned by a non-contextual model $M^{NC}$ with weights learned by a contextual model $M^C$ brings back contextual information in the decision-making process of $M^{NC}$. The degree to which contextual information is involved in the decision process is gauged in terms of accuracy improvement achieved by model $M^{NC}$ using contextual weights ($M_C^{NC}$) over that achieved by the same model using non-contextual weights ($M_{NC}^{NC}$).

The experimental setting of test WP2 implies substituting the contextual layers of some deep architecture with non-contextual (i.e., linear) layers. The resulting non-contextual attention model $M^{NC}$ is evaluated for accuracy on a test set twice: (i) using weights learned by the non-contextual model ($M_{NC}^{NC}$) and (ii) using weights derived from a contextual saliency model (in the original authors’ scenario, an LSTM with attention mechanism) which underwent a standalone training procedure ($M_C^{NC}$). If the accuracy achieved by the model $M_C^{NC}$ outperforms both that achieved by the model $M_{NC}^{NC}$ and the baseline established by $M_U^C$ in WP1, the test is passed. In other words, the test is passed if adding contextual weights to a non-contextual model improves the accuracy and leads to better performance than having uniform weights in a contextual model. In this case, the authors assert that attention is transferring contextual information to a non-contextual model. Such a claim reinforces their thesis that attention weights are not randomly learned but instead they embed complementary contextual information about the relationship among input components, being a key part of the decision process.

\section{Method}
\label{sec:method}

Previous assessments for faithfulness where analysed over one single training-test split. To make our evaluations more robust, our results involve a four-folds cross validation of the models on different training-test splits of the TCC dataset \cite{qian2020a} (further described in the appendix). Thus, we report our results in terms of average and standard deviation over such splits. Then, we perform a formal statistical analysis of the results for one leading metric, the Mean Angular Error (MAE), which we deem an appropriate concise representation of the central tendency of the angular error for the predicted illuminant (further details are provided in the appendix). For WP1, for each of the models under examination, we compared a version $M_C^C$ using learned saliency and a version $M_U^C$ using frozen random uniform weights via t-tests and adjusted the p-values using the Benjamini-Hochberg method that accounts for multiple comparisons. For WP2, we performed ANOVA tests. We fixed the MAE as the dependent variable and used as factors the saliency dimension (i.e., spatial, temporal, spatiotemporal), saliency type (i.e, attention, confidence, and both combined) and the type of weights being used for the inference (i.e., random uniform, contextual, non-contextual). The results of the ANOVAs were post-processed via the Tukey-HSD method. In both the t-tests and the ANOVAs, we measured the effect sizes using Cohen’s d value. Such a formal statistical approach provides a principled criterion to inform the decision on whether tests WP1 and WP2 were passed or failed.

Furthermore, one should note that when comparing models with the same architecture but different saliency weights, Wiegreffe \& Pinter \cite{wiegreffe2019attention} only account for the resulting accuracy, overlooking the relationship between the saliency masks involved in the test. We argue that the low difference in accuracy achieved by the two models by itself does not mean that saliency weights are not involved in the decision process. It may be that a low difference in accuracy is due to a low difference in the saliency weights used by the two models. To ascertain this possibility, we need to look at saliency weights’ divergence. In our TCC scenario, we measure the divergence between sets of saliency weights using different metrics for the spatial and temporal dimensions. When accounting for the spatiotemporal dimension, we sum the spatial and temporal metrics. We use the Jensen-Shannon Divergence (JSD) metric to account for the divergence between temporal saliency distributions. This metric was used extensively in previous work to compare probability distributions \cite{jain2019attention,serrano2019is,wiegreffe2019attention}. On the other hand, despite being scaled in $[0, 1]$, spatial saliency is not a probability distribution as it does not sum to one. Consequently, we decide to apply other measures which draw from research on comparing gray-scale images. Specifically, we choose to measure spatial divergence as the sum of binary cross-entropy, structural similarity index \cite{1284395}, and intersection over union, accounting for divergence at pixel-, patch- and feature-map-level, respectively.

The relationship between saliency weights divergence and accuracy should be looked at only when there is no statistically significant difference in accuracy between the two models, to understand whether this is due to saliency not being involved in the decision process (and thus not being faithful) or to some other reason. For a given combination of dataset and model configuration, three mutually exclusive scenarios can occur. We may observe: (i) a high difference in accuracy regardless of the divergence between sets of saliency weights, (ii) a low difference in accuracy with a high divergence between sets of saliency weights, or (iii) both a low difference in accuracy and low saliency divergence. In cases (i) and (ii), we confirm that the lack of a significant difference must be due to the saliency weights not being involved in the decision-making process. In case (iii), we cannot tell whether the test is passed or failed, and further analysis is needed. The rationale behind this scenario could either be saliency not being faithful, the model overfitting, or the occurrence of a ceiling effect in accuracy due to the task being “too easy”.

\section{Results}
\label{sec:results}

In this section, we discuss how our saliency-augmented models compare to the baseline in terms of accuracy. Then, we look at the faithfulness of the generated saliency maps. We apply the two selected tests for faithfulness (WP1 and WP2) to saliency of three types and across three dimensions of a saliency-augmented CNN+LSTM architecture. The types of saliency are Confidence (C), Attention (A), and both combined (CA). The dimensions are Spatial (S), Temporal (T) and Spatiotemporal (ST).

\subsection{Accuracy}

While making a neural model more transparent by applying some modification to its architecture, we would like its accuracy to remain unaltered (or possibly to increase). With this premise in mind, we analyze the impact on accuracy of augmenting a CNN+LSTM architecture with a saliency mechanism. More specifically, we look into how our nine proposed saliency models compare in terms of the MAE to a baseline CNN+LSTM devoid of saliency mechanisms. Despite all models performing worse than the baseline in terms of sheer numbers, this trend did not prove significant running t-tests. The small effect sizes for attention spatiotemporal, attention temporal and confidence temporal (A-ST, A-T and C-T) indicate that these models are likely to be equivalent to the baseline in terms of accuracy. 

\subsection{Test WP1}

\begin{figure*}[ht]
\centering
\includegraphics[width=.78\textwidth]{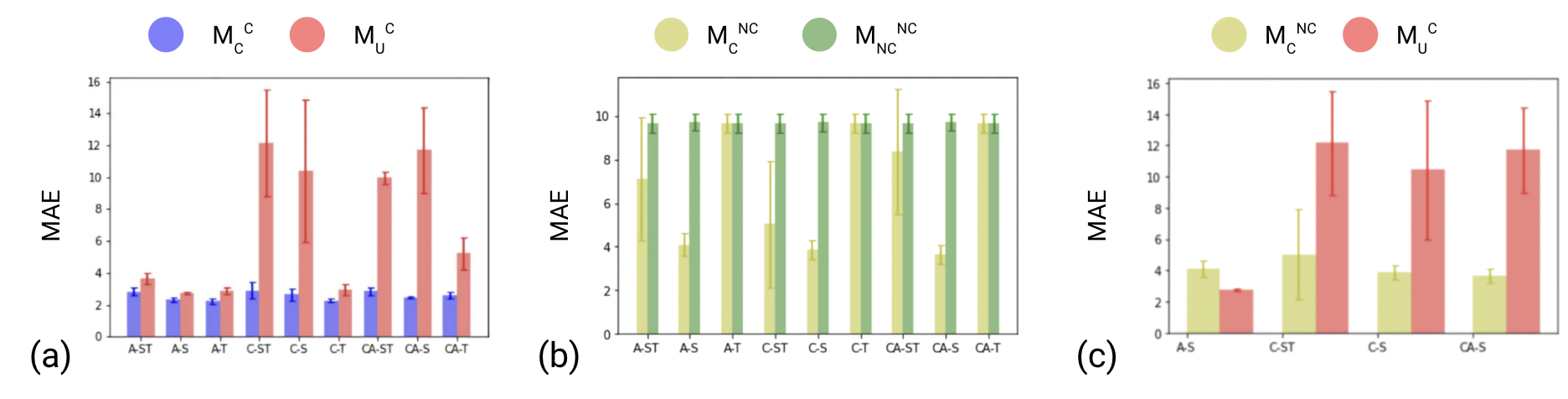}
\caption{Plot of MAE values for Test WP1 (a) and Test WP2, comparison (i) (b) and (ii) (c).}
\label{fig:results}
\end{figure*}

Figure \ref{fig:results}a compares the MAE achieved by models using either random uniformly distributed saliency weights ($M_U^C$) or saliency weights derived from learned model parameters ($M_C^C$). In terms of sheer numbers, we observe that the error achieved by models using random uniformly distributed weights is always higher than the error achieved by models using weights derived from learned model parameters. This trend is particularly accentuated when spatial confidence is involved in the evaluation. Running t-tests followed by Benjamini-Hochberg adjustments for multiple comparisons confirms that the trend is statistically significant (p-value $< 0.05$) for each of the examined configurations. Moreover, the corresponding effect sizes are very large (Cohen’s d $> 1$). Thus, all our nine proposed models pass test WP1. This means that the learned saliency scores carry some degree of information that is useful to the model for achieving accurate predictions, in that model accuracy is sensitive to manipulation of the saliency distribution. 

\subsection{Test WP2}

We remark that test WP2 is passed if the MAE achieved by the non-contextual model using saliency weights derived from a contextual model ($M_C^{NC}$) is lower than (i) the MAE achieved by the non-contextual model using weights derived from its own learned parameters ($M_{NC}^{NC}$), and (ii) the MAE achieved by the contextual model using frozen uniformly distributed saliency weights ($M_U^C$). We performed these two comparisons for each of the nine proposed saliency models as the contextual model of reference. Figure \ref{fig:results}b presents the MAE values for comparison (i) for the considered saliency types and dimensions with respect to the type of weights the model makes use of.

Looking at the bar chart, we observe that (i) holds for attention spatial, confidence spatial, confidence and attention combined spatial, and confidence spatiotemporal (A-S, C-S, CA-S, C-ST). For the other configurations, we need to check if the lack of a significant difference could be due to low divergence in the saliency masks generated by the non-contextual model using learned saliency weights ($M_{NC}^{NC}$) and the non-contextual model using saliency weights imposed from the contextual model ($M_C^{NC}$). To this purpose, we look at the relationships between accuracy and the generated saliency masks for both the temporal and spatiotemporal models, and interpret them as discussed in §\ref{sec:method}. For all of the considered models, saliency divergence is high (i.e., $Div_{temp} > 0.7,~Div_{spat} > 125$), which suggests that the low difference in accuracy is not due to saliency weights being very similar, but rather to them not being involved in the decision-making process of the model. Therefore, we do not consider these models when analysing comparison (ii). 

As shown by the plot in Figure \ref{fig:results}c, comparison (ii) holds for confidence spatiotemporal, confidence spatial and confidence and attention combined spatial (C-ST, C-S, CA-S). Thus, these models pass test WP2. On the other hand, comparison (ii) does not hold for attention spatial (A-S). In this case, the contextual architecture of the model has more impact on accuracy than the saliency mechanism.

\begin{figure}[t]
\centering
\includegraphics[width=.4\textwidth]{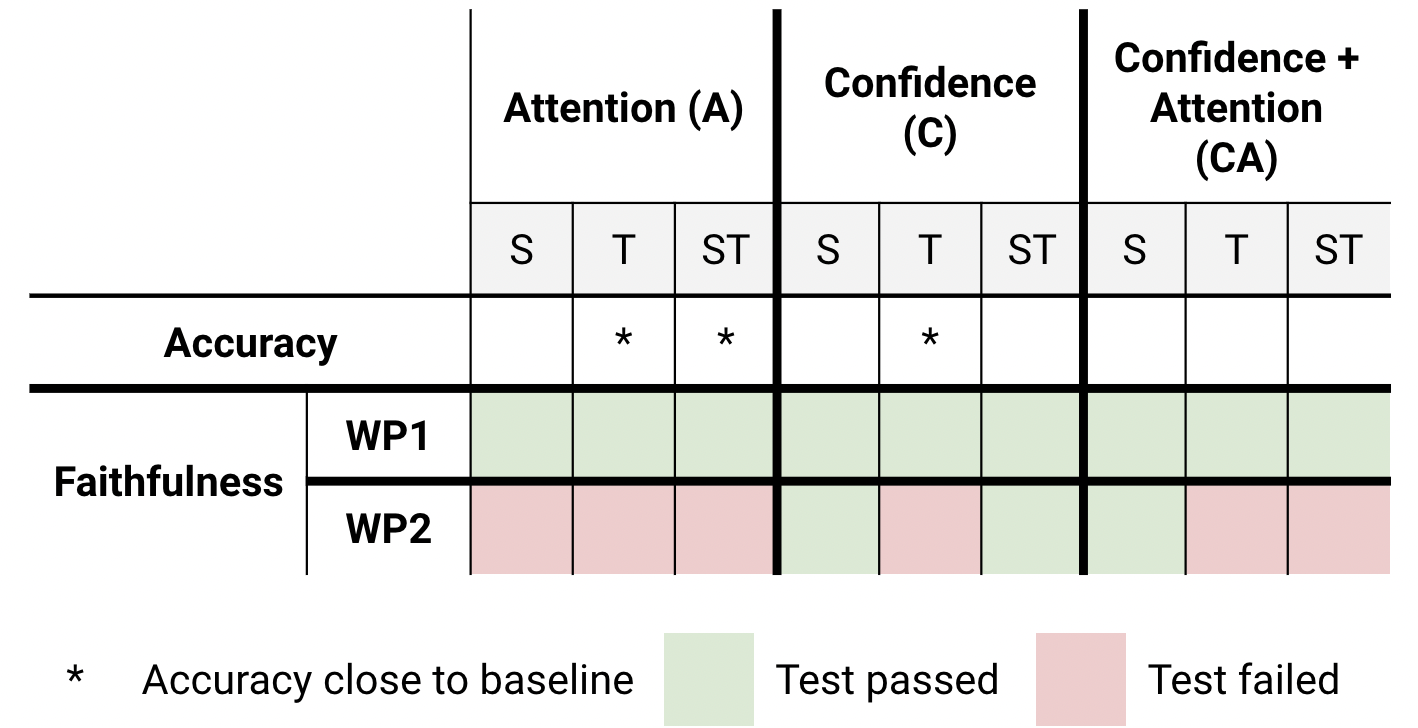}
\caption{Summary table of the results of tests WP1 and WP2.}
\label{fig:summary-table}
\end{figure}

\subsection{Discussion}

Figure \ref{fig:summary-table} summarizes the results of the analysis of accuracy and of the two tests for faithfulness for our nine models. 

First, we note that the temporal dimension is present in all three the top-performing model configurations, suggesting that this facet of saliency is important for accuracy. However, no model achieves faithfulness when leveraging the temporal dimension only. An intuitive justification for this phenomenon stems from observing that temporal saliency tends to focus on very few frames in a sequence. This means that a lot of possibly relevant spatial information is discarded. Thus, a model might learn not to actively use temporal saliency in its decision-making process to preserve accuracy. 

Second, we observe that spatial confidence is present in all of the configurations that pass the two assessments, which suggests that this saliency dimension and type help to achieve faithfulness. This might be due to confidence scores being jointly learned with the other feature maps and thus more strictly connected to the inner decision-making process of the model. On the other hand, attention configurations never achieve faithfulness. The main crucial difference between attention and confidence is that the former is learned by a separate ad hoc convolutional module while the latter is learned as an additional feature map. As a result, the attention model is more complex (i.e., has a larger number of trainable parameters, $\sim$x3). It might be that attention networks get sufficiently complex to achieve high accuracy while ignoring saliency in their decision-making process. That is, there could be a trade-off between complexity and faithfulness. In other words, a model could be complex enough to solve the task effectively and introducing further parameters may result in a ceiling effect. In this case, the model would still be able to fulfil the target task without fully leveraging its complexity and thus regardless of the learned saliency weights.

\section{Conclusions and Future Work}

In this work, we examined explanations generated by in-model saliency methods for DL models in the unexplored context of video-based illuminant estimation. In particular, we assessed the property of faithfulness, crucial to enable user tasks such as debugging and improving model architectures, for two types of saliency (i.e., attention and confidence) and across three dimensions (i.e., spatial, temporal and spatiotemporal). Our assessment was performed via two tests from previous NLP literature, whose methodology we improve by introducing a statistical analysis of the results and a study of the divergence of the sets of saliency weights involved. The results of the tests suggest that only spatial and spatiotemporal confidence lead to faithful explanations. All the considered attention models fail to achieve faithfulness, seemingly confirming past research discrediting it as a reliable form of explanation. We also analyzed the models in terms of accuracy and found that temporal models seem to lead to better performance.

Now that we have assessed the faithfulness of in-model saliency, the next step of our research will involve trying to understand if the generated explanations are meaningful to the end-users. We plan to do this by clustering the salient parts of the inputs and running a user study to test whether the semantic of the clusters align to human expertise on colour constancy.

Lastly, our work is not devoid of limitations. Most importantly, our experiments are circumscribed to a single task and a single dataset, which makes it hard to generalize the results to other domains. As part of our future work, we plan to gather all the datasets used in the identified tests by Jain \& Wallace, Wiegreffe \& Pinter, Serrano \& Smith and Adebayo et al. and apply consistently these assessments on all available data, both for TCC and for the NLP tasks studied by previous works. Furthermore, despite advocating for rigour in our analysis, we could not specify a clear threshold to define when not passing a test is due to a lack of divergence in saliency weights. We aim to devise such a threshold in future work. Despite these limitations, our work still serves as a first attempt on assessing the faithfulness of explanations generated by in-model saliency for the TCC task, as well as fixing general methodological issues of the existing tests.

\bibliographystyle{unsrt}
\bibliography{ijcai22}

\clearpage
\appendix

\section{Adaptation of the Tests for Faithfulness}

Tests WP1 and WP2 were originally proposed for experimenting on binary text classification using a vanilla LSTM architecture. Thus, they need to be adapted to our target TCC task and CNN+LSTM design.

WP1 is a model and task agnostic test that requires minor adaptations. We generated the required random uniform distribution specified by the original methodology at inference time both for the spatial and temporal saliency dimensions, independently, using the same strategy for attention and confidence (i.e., the utility function offered by the PyTorch framework for DL v1.9.0). The uniformly distributed weights were then used to replace those learned by the model.  In the spatial and temporal cases, we froze to a random uniform distribution the saliency weights of the corresponding module leaving the saliency weights of the other module unaltered, while we tweaked the saliency weights of both modules at once for the spatiotemporal scenario.

Concerning test WP2, we selectively replaced the convolutional and recurrent layers of the CNN+LSTM architecture based on which dimension, among spatial, temporal, and spatiotemporal, we were examining. This is done to prevent the network to access information about the relationship between input components, such as pixels in the frames and timesteps in the sequences. In the case of the evaluation of spatial saliency, we applied one linear layer in place of the convolutional component while keeping unaltered the recurrent component, and vice versa for the evaluation of temporal saliency. When looking at spatiotemporal saliency, we replaced both the convolutional and recurrent layers with linear layers.

\section{Details of the Experimental Setup}

We evaluated our models on the TCC dataset \cite{qian2020a}, which is the largest and most realistic dataset available for TCC\footnote{Of the other existing datasets for TCC, \cite{prinet2013illuminant} is very small, \cite{yoo2019dichromatic} was specifically designed for experiments on AC bulb illumination and \cite{ciurea2003a} features very low-resolution images which are not in line with the standards of modern consumer photography.}. It consists of 600 real-world videos recorded with a high-resolution mobile phone camera shooting 1824 x 1368 sized pictures. The length of these videos ranges from 3 to 17 frames (7.3 on average, the median is 7.0 and mode is 8.5). Ground truth information is present only for the last frame in each video (i.e., the shot frame), and was collected using a grey surface calibration target. Despite TCC being the largest dataset available, the number of sequences is relatively small for training a DL model effectively. Therefore, we perform data augmentation according to the same procedure used in \cite{qian2020a} to train TCCNet. Namely, the frames in the sequences were randomly rotated by an amount in the range [-30, +30] degrees and cropped to a proportion in the range of [0.8, 1.0] on the shorter dimension. The generated patches were flipped horizontally with a probability of 0.5. Data augmentation is achieved by dynamically applying these transformations at training time to each batch of sequences, which is the standard practice in PyTorch, our DL framework of reference.

All models have been trained using a mix of Tesla P100 and NVidia GeForce GTX 1080 Ti GPUs from local lab equipment and cloud services taking about 24 hours for a model to complete a single learning procedure. The training of the saliency models was performed for 500 epochs using the RMSprop optimizer with batch size 1 and learning rate initially set to $3e^{-5}$. We opted for hidden size equal 128 and kernel size equal 5, as suggested by the ablation study of TCCNet in \cite{qian2020a}. The SqueezeNet backbone was initialised with the weights pretrained on ImageNet \cite{deng2009imagenet} provided by PyTorch\footnote{The pretrained models offered by PyTorch are available at https://pytorch.org/docs/stable/torchvision/models.html}. The error $\epsilon$ between the illuminant $\hat{c}$ estimated by the saliency models and the ground truth $c_{gt}$ has been computed using the angular error, a measure used in many works on computational colour constancy and reported in Formula \ref{eq:angular-error}.

\begin{equation}
\label{eq:angular-error}
\epsilon_{\hat{c}, c_{gt}} = arccos(\frac{\hat{c} \cdot c_{gt}}{||\hat{c}|| \cdot ||c_{gt}||})
\end{equation}

As in previous works on computational color constancy, the metrics we selected to evaluate model performance in terms of accuracy provide insights into the distribution of the angular errors across the test items. These metrics include the MAE, the median and the trimean (defined as the weighted average of the median and upper and lower quartiles) across the test set, indicating how the models performed on average and accounting for outliers. We also report the MEA on the best 25th and worst 25th and 5th percentiles, showing how the model performed on easy, hard, and very hard inputs, respectively.

\section{Extended Results of the Experiments}

\begin{table*}[h!]
    \centering
    \resizebox{\textwidth}{!}{
    \begin{tabular}{cccccccccccccc}
    \multirow{2}{*}{ \textbf{Model} } & \multirow{2}{*}{ \textbf{Saliency} } & \multicolumn{2}{c}{\textbf{Mean}} & \multicolumn{2}{c}{\textbf{Median}} & \multicolumn{2}{c}{\textbf{Trimean}} & \multicolumn{2}{c}{\textbf{Best 25\%}} & \multicolumn{2}{c}{\textbf{Worst 25\%}} & \multicolumn{2}{c}{\textbf{Worst 5\%}} \\
    & & Avg & Std dev & Avg & Std dev & Avg & Std dev & Avg & Std dev & Avg & Std dev & Avg & Std dev \\
    B & - & 2.28 & 0.24 & 1.50 & 0.19 & 1.72 & 0.23 & 0.46 & 0.06 & 5.32 & 0.58 & 6.60 & 1.13 \\
    \multirow{6}{*}{ A } & ST (L) & 2.83 & 0.27 & 1.87 & 0.19 & 2.13 & 0.22 & 0.45 & 0.05 & 6.78 & 0.76 & 9.08 & 1.04 \\
    & ST (R) & 3.66 & 0.34 & 2.42 & 0.24 & 2.69 & 0.27 & 0.67 & 0.10 & 8.77 & 1.02 & 11.42 & 1.28 \\
    & S (L) & 2.32 & 0.14 & 1.66 & 0.05 & 1.81 & 0.05 & 0.45 & 0.04 & 5.34 & 0.56 & 6.49 & 1.16 \\
    & S (R) & 2.74 & 0.06 & 1.99 & 0.08 & 2.20 & 0.09 & 0.58 & 0.04 & 6.08 & 0.18 & 7.22 & 0.22 \\
    & T (L) & 2.22 & 0.18 & 1.26 & 0.07 & 1.50 & 0.07 & 0.28 & 0.04 & 5.79 & 0.58 & 7.43 & 1.35 \\
    & T (R) & 2.90 & 0.21 & 1.87 & 0.06 & 2.14 & 0.10 & 0.53 & 0.11 & 6.92 & 0.56 & 8.85 & 1.08 \\
    \multirow{6}{*}{ C } & ST (L) & 2.90 & 0.52 & 2.24 & 0.50 & 2.35 & 0.51 & 0.64 & 0.15 & 6.44 & 1.05 & 8.00 & 1.19 \\
    & ST (R) & 12.16 & 3.34 & 11.06 & 2.59 & 11.33 & 2.86 & 2.94 & 1.32 & 23.34 & 6.67 & 25.80 & 8.61 \\
    & S (L) & 2.64 & 0.37 & 1.98 & 0.37 & 2.11 & 0.36 & 0.60 & 0.13 & 5.85 & 0.70 & 6.83 & 1.22 \\
    & S (R) & 10.43 & 4.47 & 9.76 & 5.31 & 10.00 & 5.10 & 3.93 & 2.32 & 18.00 & 5.85 & 20.13 & 5.91 \\
    & T (L) & 2.28 & 0.08 & 1.64 & 0.11 & 1.81 & 0.07 & 0.49 & 0.06 & 5.07 & 0.30 & 6.37 & 0.10 \\
    & T (R) & 2.94 & 0.38 & 2.15 & 0.28 & 2.31 & 0.29 & 0.66 & 0.11 & 6.57 & 0.99 & 8.29 & 1.37 \\
    \multirow{6}{*}{ CA } & ST (L) & 2.84 & 0.27 & 2.05 & 0.21 & 2.27 & 0.22 & 0.59 & 0.12 & 6.44 & 0.65 & 7.94 & 1.52 \\
    & ST (R) & 9.97 & 0.38 & 9.08 & 0.81 & 9.49 & 0.41 & 3.08 & 0.13 & 18.36 & 1.37 & 20.38 & 1.86 \\
    & S (L) & 2.45 & 0.07 & 1.71 & 0.09 & 1.89 & 0.06 & 0.47 & 0.07 & 5.66 & 0.30 & 6.99 & 0.21 \\
    & S (R) & 11.70 & 2.71 & 9.68 & 2.65 & 10.57 & 2.96 & 2.72 & 0.78 & 23.11 & 4.41 & 25.83 & 4.22 \\
    & T (L) & 2.60 & 0.22 & 1.77 & 0.22 & 1.98 & 0.22 & 0.52 & 0.10 & 6.04 & 0.43 & 7.62 & 0.57 \\
    & T (R) & 5.23 & 1.00 & 3.75 & 0.91 & 4.13 & 1.03 & 0.84 & 0.07 & 12.04 & 2.14 & 14.77 & 2.93 \\
    \end{tabular}
    }
    \caption{Test WP1. Metrics with respect to the angular error for models using random uniformly distributed (R) versus learned (L) saliency weights.}
    \label{tab:exp-wp1}
\end{table*}

\begin{table*}[h!]
    \centering
    \resizebox{\textwidth}{!}{
    \begin{tabular}{cccccccccccccc}
    \multirow{2}{*}{ \textbf{Saliency Type} } & \multirow{2}{*}{ \textbf{Saliency} } & \multicolumn{2}{c}{\textbf{Mean}} & \multicolumn{2}{c}{\textbf{Median}} & \multicolumn{2}{c}{\textbf{Trimean}} & \multicolumn{2}{c}{\textbf{Best 25\%}} & \multicolumn{2}{c}{\textbf{Worst 25\%}} & \multicolumn{2}{c}{\textbf{Worst 5\%}} \\
    & & Avg & Std Dev & Avg & Std Dev & Avg & Std Dev & Avg & Std Dev & Avg & Std Dev & Avg & Std Dev \\
    \multirow{3}{*}{ Baseline } & ST & 9.89 & 0.44 & 9.74 & 0.30 & 9.53 & 0.55 & 2.63 & 0.64 & 17.86 & 0.75 & 19.48 & 1.18 \\
    & S & 4.10 & 0.51 & 9.25 & 9.31 & 2.82 & 17.72 & 19.44 & 9.80 & 9.23 & 9.25 & 2.79 & 17.81 \\
    & T & 9.91 & 0.45 & 9.99 & 0.55 & 9.64 & 0.57 & 2.69 & 0.46 & 17.75 & 0.69 & 19.29 & 0.77 \\
    \multirow{3}{*}{ Learned } & ST & 9.64 & 0.42 & 9.03 & 0.49 & 9.10 & 0.51 & 2.05 & 0.43 & 18.20 & 0.82 & 20.47 & 0.67 \\
    & S & 9.70 & 0.40 & 9.35 & 0.51 & 9.21 & 0.46 & 2.56 & 0.23 & 17.99 & 0.86 & 19.64 & 0.45 \\
    & T & 9.64 & 0.43 & 8.96 & 0.56 & 9.07 & 0.50 & 2.31 & 0.58 & 18.16 & 0.62 & 20.37 & 0.85 \\
    \multirow{3}{*}{ Contextual } & ST & 7.12 & 2.81 & 6.21 & 3.24 & 6.31 & 3.08 & 1.65 & 0.84 & 14.42 & 4.20 & 17.24 & 3.60 \\
    & S & 4.10 & 0.51 & 3.18 & 0.25 & 3.40 & 0.35 & 1.05 & 0.08 & 8.79 & 1.40 & 11.01 & 1.72 \\
    & T & 9.65 & 0.43 & 9.30 & 0.68 & 9.17 & 0.61 & 2.39 & 0.31 & 18.06 & 0.73 & 20.49 & 0.71 \\
    \end{tabular}
    }
    \caption{Test WP2. Metrics with respect to the angular error for models using \textit{attention}.}
    \label{tab:exp-wp2-attention}
\end{table*}

\begin{table*}[h!]
    \centering
    \resizebox{\textwidth}{!}{
    \begin{tabular}{cccccccccccccc}
    \multirow{2}{*}{ \textbf{Saliency Type} } & \multirow{2}{*}{ \textbf{Saliency} } & \multicolumn{2}{c}{\textbf{Mean}} & \multicolumn{2}{c}{\textbf{Median}} & \multicolumn{2}{c}{\textbf{Trimean}} & \multicolumn{2}{c}{\textbf{Best 25\%}} & \multicolumn{2}{c}{\textbf{Worst 25\%}} & \multicolumn{2}{c}{\textbf{Worst 5\%}} \\
    & & Avg & Std Dev & Avg & Std Dev & Avg & Std Dev & Avg & Std Dev & Avg & Std Dev & Avg & Std Dev \\
    \multirow{3}{*}{ Baseline } & ST & 9.89 & 0.44 & 9.70 & 0.36 & 9.52 & 0.57 & 2.65 & 0.64 & 17.84 & 0.78 & 19.44 & 1.19 \\
    & S & 19.35 & 9.83 & 9.49 & 9.44 & 2.89 & 17.65 & 19.47 & 0.22 & 8.22 & 0.91 & 9.59 & 1.01 \\
    & T & 9.91 & 0.45 & 9.98 & 0.57 & 9.63 & 0.58 & 2.70 & 0.45 & 17.74 & 0.70 & 19.28 & 0.77 \\
    \multirow{3}{*}{ Learned } & ST & 9.64 & 0.42 & 9.10 & 0.44 & 9.12 & 0.50 & 2.42 & 0.50 & 18.09 & 0.78 & 20.51 & 0.74 \\
    & S & 9.69 & 0.42 & 9.20 & 0.28 & 9.26 & 0.42 & 2.66 & 0.41 & 18.01 & 0.52 & 20.32 & 0.56 \\
    & T & 9.65 & 0.42 & 9.01 & 0.53 & 9.06 & 0.50 & 2.34 & 0.56 & 18.10 & 0.67 & 20.33 & 1.03 \\
    \multirow{3}{*}{ Contextual } & ST & 5.03 & 2.89 & 4.14 & 3.11 & 4.33 & 3.02 & 1.09 & 0.80 & 10.51 & 4.58 & 13.09 & 5.09 \\
    & S & 3.86 & 0.43 & 2.99 & 0.44 & 3.20 & 0.37 & 0.96 & 0.22 & 8.22 & 0.91 & 9.59 & 1.01 \\
    & T & 9.65 & 0.42 & 9.30 & 0.66 & 9.19 & 0.59 & 2.35 & 0.40 & 18.12 & 0.91 & 20.43 & 0.74 \\
    \end{tabular}
    }
    \caption{Test WP2. Metrics with respect to the angular error for models using \textit{confidence}.}
    \label{tab:exp-wp2-confidence}
\end{table*}

\begin{table*}[h!]
    \centering
    \resizebox{\textwidth}{!}{
    \begin{tabular}{ccccccccccccc}
    \multirow{2}{*}{ \textbf{Saliency} } & \multicolumn{2}{c}{\textbf{Mean}} & \multicolumn{2}{c}{\textbf{Median}} & \multicolumn{2}{c}{\textbf{Trimean}} & \multicolumn{2}{c}{\textbf{Best 25\%}} & \multicolumn{2}{c}{\textbf{Worst 25\%}} & \multicolumn{2}{c}{\textbf{Worst 5\%}} \\
    & Avg & Std Dev & Avg & Std Dev & Avg & Std Dev & Avg & Std Dev & Avg & Std Dev & Avg & Std Dev \\
    ST & 9.89 & 0.44 & 9.71 & 0.39 & 9.53 & 0.57 & 2.67 & 0.63 & 17.82 & 0.79 & 19.38 & 1.19 \\
    S & 9.83 & 0.49 & 9.49 & 0.53 & 9.44 & 0.70 & 2.89 & 0.53 & 17.65 & 0.41 & 19.47 & 0.68 \\
    T & 9.91 & 0.45 & 9.99 & 0.56 & 9.64 & 0.58 & 2.69 & 0.46 & 17.75 & 0.69 & 19.29 & 0.77 \\
    ST & 9.64 & 0.42 & 9.12 & 0.44 & 9.13 & 0.50 & 2.43 & 0.50 & 18.07 & 0.79 & 20.46 & 0.75 \\
    S & 9.69 & 0.39 & 9.16 & 0.60 & 9.17 & 0.49 & 2.60 & 0.34 & 18.04 & 0.82 & 19.76 & 0.54 \\
    T & 9.64 & 0.43 & 8.95 & 0.58 & 9.05 & 0.52 & 2.31 & 0.59 & 18.16 & 0.62 & 20.40 & 0.85 \\
    ST & 8.35 & 2.85 & 7.81 & 3.41 & 7.76 & 3.20 & 2.02 & 0.87 & 16.16 & 4.33 & 18.22 & 3.99 \\
    S & 3.64 & 0.45 & 2.73 & 0.45 & 2.97 & 0.46 & 0.78 & 0.13 & 7.97 & 0.71 & 9.73 & 0.86 \\
    T & 9.66 & 0.42 & 9.28 & 0.74 & 9.19 & 0.64 & 2.41 & 0.30 & 18.05 & 0.74 & 20.30 & 0.45 \\
    \end{tabular}
    }
    \caption{Test WP2. Metrics with respect to the angular error for models using \textit{confidence} as spatial saliency and \textit{attention} as temporal saliency}
    \label{tab:exp-wp2-confidence}
\end{table*}

\end{document}